\documentclass[letterpaper, 10 pt, conference]{ieeeconf}  

\IEEEoverridecommandlockouts                              

\usepackage{color}
\usepackage{graphicx}
\usepackage{tabularx}
\usepackage{crop}
\usepackage{multirow} 
\usepackage{caption}
\usepackage{amsmath}
\usepackage{amssymb}
\usepackage{xspace}
\usepackage{subfig}
\usepackage{booktabs}
\usepackage{dsfont}

\usepackage[export]{adjustbox}

\newcolumntype{Y}{>{\centering\arraybackslash}X}

\def\eg{\textit{e.g.}\xspace}
\def\ie{\textit{i.e.}\xspace}

\title{\LARGE \bf Kitting in the Wild through Online Domain Adaptation
}

\author{Massimiliano Mancini$^{\star 1,2}$, Hakan Karaoguz$^{\star 3}$, Elisa Ricci$^{2,4}$, Patric Jensfelt$^{3}$, Barbara Caputo$^{1,5}$
\thanks{This work was partially supported by The Swedish Foundation for Strategic Research (SSF) and its Centre for Autonomous Systems and the project FACT (H.K., P. J.) and the ERC project RoboExNovo (M. M., B.C.).}
\thanks{$^{1}$M. Mancini and B. Caputo are with Sapienza University of Rome, Rome, Italy.{\tt\small \{mancini,caputo\}@diag.uniroma1.it}}%
\thanks{$^{2}$M. Mancini and E. Ricci are with Fondazione Bruno Kessler, Trento, Italy. {\tt\small eliricci@fbk.eu}}%
\thanks{$^{3}$H. Karaoguz and P. Jensfelt are with KTH Royal Institute of Technology, Stockholm, Sweden.  {\tt\small \{hkarao,patric\}@kth.se}}
\thanks{$^{4}$E. Ricci is with University of Trento, Trento, Italy.}
\thanks{$^{5}$B. Caputo is with Italian Institute of Technology, Milan, Italy.}
\thanks{$^\star$ denotes equal contribution.}
}

\begin{document}

\begin{titlepage}
\null
\vfill
\renewcommand{\fboxsep}{10pt}
\fbox{\Large\begin{minipage}{\columnwidth}
\textbf{Disclaimer:}

This work has been accepted for publication in the Proceedings of the 2018 IEEE/RSJ International Conference on Intelligent Robots and Systems\vspace{4pt}
\newline
link:   https://www.iros2018.org/
\newline
\newline
\textbf{Copyright:} 
\newline
\copyright~2018 IEEE. Personal use of this material is permitted. Permission from IEEE must be obtained for all other uses,  in  any  current  or  future  media,  including  reprinting/  republishing  this  material  for  advertising  or promotional purposes, creating new collective works, for resale or redistribution to servers or lists, or reuse of any copyrighted component of this work in other works.
\newline
\end{minipage}}
\vfill
\clearpage
\end{titlepage}

\maketitle
\thispagestyle{empty}
\pagestyle{empty}

\begin{abstract}
Technological developments call for increasing perception and action capabilities of robots. Among other skills, vision systems that can adapt to any possible change in the working conditions are needed. Since these conditions are unpredictable, we need benchmarks which allow to assess the generalization and robustness capabilities of our visual recognition algorithms. 
In this work we focus on robotic kitting in unconstrained scenarios. As a first contribution, we present a new visual dataset for the kitting task. Differently from standard object recognition datasets, we provide images of the same objects acquired under various conditions where camera, illumination and background are changed. This novel dataset allows for testing the robustness of robot visual recognition algorithms to a series of different \emph{domain shifts} both in isolation and unified. 
Our second contribution is a novel online adaptation algorithm for deep models, based on batch-normalization layers, which allows to continuously adapt a model to the current working conditions. Differently from standard domain adaptation algorithms, it does not require any image from the target domain at training time. We benchmark the performance of the algorithm on the proposed dataset, showing its capability to fill the gap between the performances of a standard architecture and its counterpart adapted offline to the given target domain.

\end{abstract}

\section{INTRODUCTION}
\label{sec:intro}
\begin{figure}[t]
\centering
\includegraphics[width=1\columnwidth]{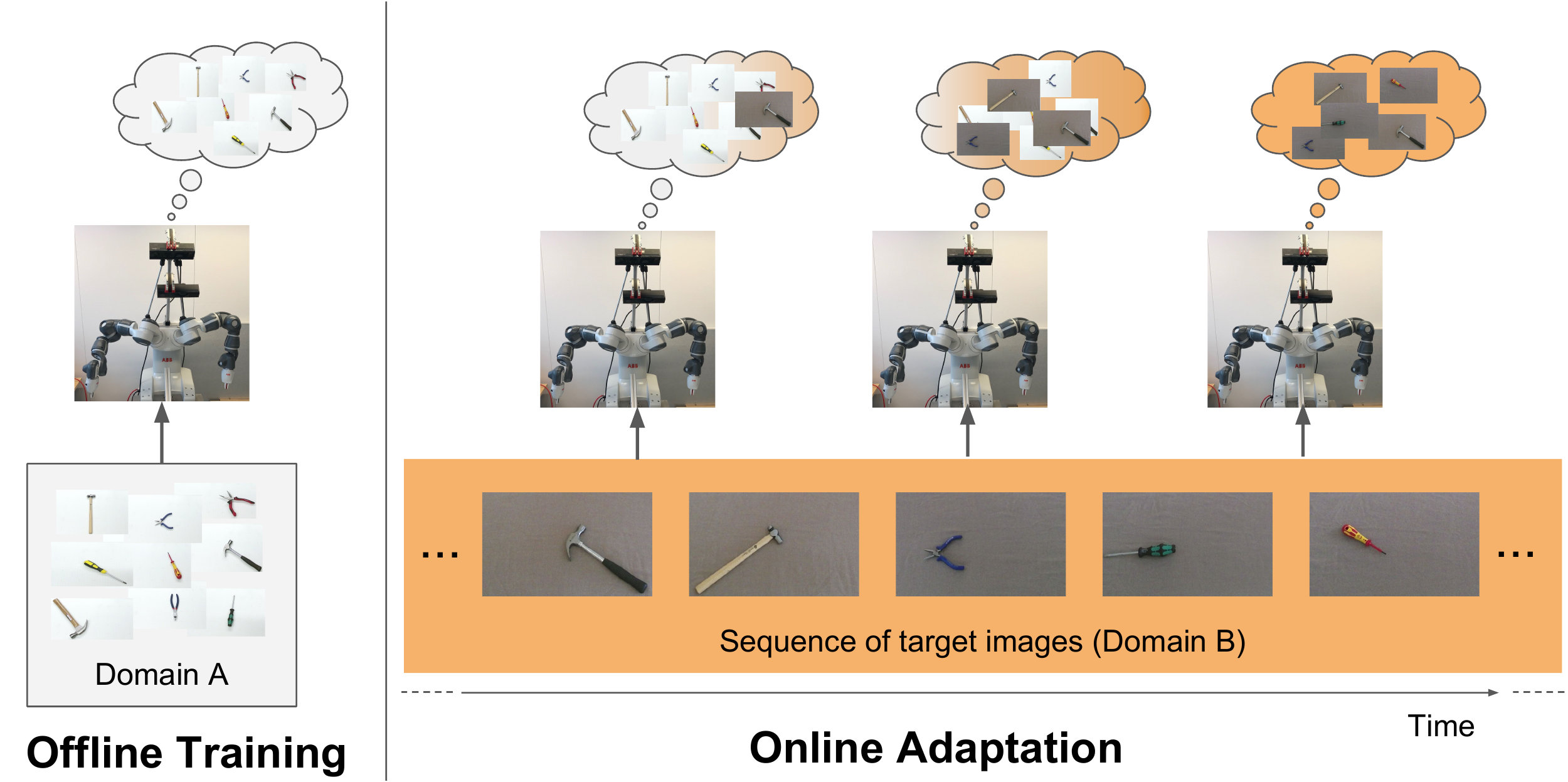}
    \caption{Our approach for performing kitting in arbitrary conditions. Given a training set. we can train a robot vision model offline. As the robot performs the task, we gradually adapt the visual model to the current working conditions, in an online fashion and without requiring target data during the offline training phase. } 
    \label{fig:teaser}
\end{figure}

Robot technology is already an integral part of the manufacturing industry. Robots are currently being used in variety of tasks such as machine loading, part inspection, bin picking, kitting, and assembly. 
In this work, we consider the task of kitting which is the process of grouping related parts such as gathering components of a personal computer (PC) into one bin for assembly~\cite{banerjee2015ontology}. 
The kitting task requires the recognition of the parts in the environment, the ability to pick objects from the bins and placing them at the correct location \cite{holz2015skill}. All of these subtasks are very challenging on their own but the recognition of the parts is crucial for the robot to sequentially perform the other subtasks. Already in today's factory settings, object recognition tasks possess challenges such as environmental effects (illumination, viewpoint, etc), varying object material properties and cluttered scenes \cite{liu2012fast}. In order to simplify the recognition task, some approaches use machine vision in rather isolated settings for decreasing the environmental variability \cite{Schyja12}. Liu et. al \cite{liu2012fast} proposed a specially designed camera system and estimation based on 3D CAD models to robustly detect and verify the type and the pose of the object. Kaiba et. al. \cite{kaipa2015resolving} proposed an interactive method where a remote human operator resolves ambiguities in the perception system. Unfortunately, none of the above methods are generic enough to be applied in a truly unconstrained setting. In this paper we are primarily concerned with solving the object recognition problem for kitting using vision in the wild, i.e. in non-isolated settings exhibiting large variations. Right now, most of the robots in manufacturing industry are operating in isolation, primarily because of safety concerns. However, many future scenarios have robots and humans working closer together, moving robots into new areas of applications, beyond mass production and preprogrammed behavior. For this to happen, not only safety, but perception will be a major challenge.

In the last years deep neural networks \cite{krizhevsky2012imagenet} 
have become the new dominant learning paradigm in visual recognition,
establishing the new state of the art in various visual tasks such as object classification \cite{he2016deep} and object detection \cite{ren2017faster}. Similarly deep architectures have been applied on real robots \cite{karaoguz2017human,Young17}, leading to significant improvements on a variety of robot vision tasks \cite{pasquale2016object,hoffman2016icra,zaki2016convolutional}. 
One known challenge with DNNs is that they are data hungry. This is particularly problematic for robotic scenarios where the data collection process can be costly or even unfeasible, thus the amount of training data is limited. Some authors proposed to overcome this issue by leveraging over synthetic data \cite{carlucci2016deep}, but while this approach seems promising for depth data, it is questionable if it will work on RGB images.
Domain Adaptation (DA, \cite{DBLP:journals/corr/Csurka17}) attempts to circumvent this issue  by adapting a model from a given domain for which sufficient training data is available, denoted as \textit{source} domain, to a domain for which few or no labeled data are available, called the \textit{target} domain.  Despite the remarkable performances achieved by DA algorithms in computer vision \cite{li1603revisiting,carlucci2017autodial}, and their growing popularity in robot vision \cite{angeletti2017icra} they require the presence of images from the target domain in advance during training. This is a huge limitation due to the likely unpredictable conditions of the environment in which a robot is employed.

This paper attempts to advance the state of the art in kitting in realistic deployments with two contributions. 
First, we propose a novel kitting dataset which contains images of objects taken under varying illumination, viewpoint and background conditions from a robotic platform. This dataset, that upon acceptance of the paper will be made publicly available through a dedicated website, provides the community with a novel tool for studying the robustness of robot vision algorithms to drastic changes in the appearance of the input images, and assess progress in the field. We are not aware of existing, publicly available kitting databases covering this range of visual variability.  

Second, we propose a novel approach for achieving \textit{online} adaptation of a deep model. Differently from classical DA approaches, our algorithm can adapt a deep model to any target domain on the fly, without requiring any target domain data before-hand. We benchmark the performance of our algorithm on the presented dataset, showing how this model is able to produce large improvements on the target domain performances compared to the base architecture trained on the source domain, and matching what would be achieved by having all data from the target available beforehand.



The outline for the rest of the paper is as follows. In Section~\ref{sec:related-work} we give an overview of related work. In Section~\ref{sec:dataset} we present the new dataset, describing the collection process and the data contained. 
Section~\ref{sec:method} describes our online DA method and Section~\ref{sec:experiments} presents the result of our evaluation. Finally, Section~\ref{sec:conclusions} gives conclusions and outlines avenues for future research.

\section{Related work}
\subsection{Kitting task}
Robotic kitting and bin picking are similar and well known problems. Several methods have been proposed to solve these problems by either using specialized setups \cite{liu2012fast}, high-level frameworks \cite{holz2015skill,Schyja12} or using human-robot collaboration \cite{banerjee2015ontology,kaipa2015resolving}. 
Liu et al. in \cite{liu2012fast} use a customized camera for extracting edges of the objects in a bin and then using shape-matching to detect objects and estimating their poses. The proposed algorithm is computationally efficient so that it can be deployed in real robot scenario. 
Holz et al. in \cite{holz2015skill} propose a high-level framework that is composed of individual modules such as object detection, planning etc. to automatically perform kitting task in real world scenarios. Similarly in \cite{Schyja12} a high-level framework that combines virtual and real setups is proposed. Virtual setup helps to optimize the actual system without any risk of collisions. Therefore the real system can be setup in more economical way with less number of actual trials. Banerjee et al. in \cite{banerjee2015ontology} employ human-robot interaction for performing the kitting task. They first present a common ontology for representing all the required subtasks for kitting. Then these subtasks are optimally partitioned between the robots and humans to complete the whole task faster and with less failures. Kaipa et al. in \cite{kaipa2015resolving} present a method where a remote human operator assists the robot for selecting the object when the automated perception system fails.

\subsection{Online Domain Adaptation}
Recent years have witnessed great advances in domain adaptation both in computer \cite{long2015learning,carlucci2017autodial,li1603revisiting} and robot vision \cite{angeletti2017icra,hoffman2016icra,mancini18}. Adapting a model from one domain to another requires to bridge the gap between the two different distributions generating the data of different domains. In deep learning architectures this is usually achieved by minimizing the difference between the features produced by images of different domains, \eg through domain confusion losses or by minimizing discrepancy measures \cite{long2015learning,tzeng2015simultaneous,ganin2014unsupervised}. Recently, it has been shown how batch-normalization (BN) layers \cite{ioffe2015batch} can be employed to match the source and target data distributions by applying different BN layers for each domain \cite{li1603revisiting,carlucci2017just,carlucci2017autodial}.  Our method develops from this last research trend, with BN statistics adapted to the images of the \textit{experienced} novel domain. Differently from \cite{li1603revisiting,carlucci2017just,carlucci2017autodial} it does not require any target domain data during training. 


We would like to remark that, despite its simplicity, our approach is the first deep domain adaptation method that operates in an online setting, without requiring any prior about the target domain.

Another related research trend is domain generalization (DG). DG frameworks \cite{li2017deeper,mancini18} aim at generalizing a model from multiple, given, source domains, to any target domain, with no data of the target domain available at training time. Differently from these techniques, we assume to have only one source domain during training, without the need of multiple data acquisition and labeling processes.
\label{sec:related-work}

\section{KTH Handtool Dataset}
\label{sec:dataset}

\begin{figure}[htb!]
      \centering
      \includegraphics[width=0.8\columnwidth]{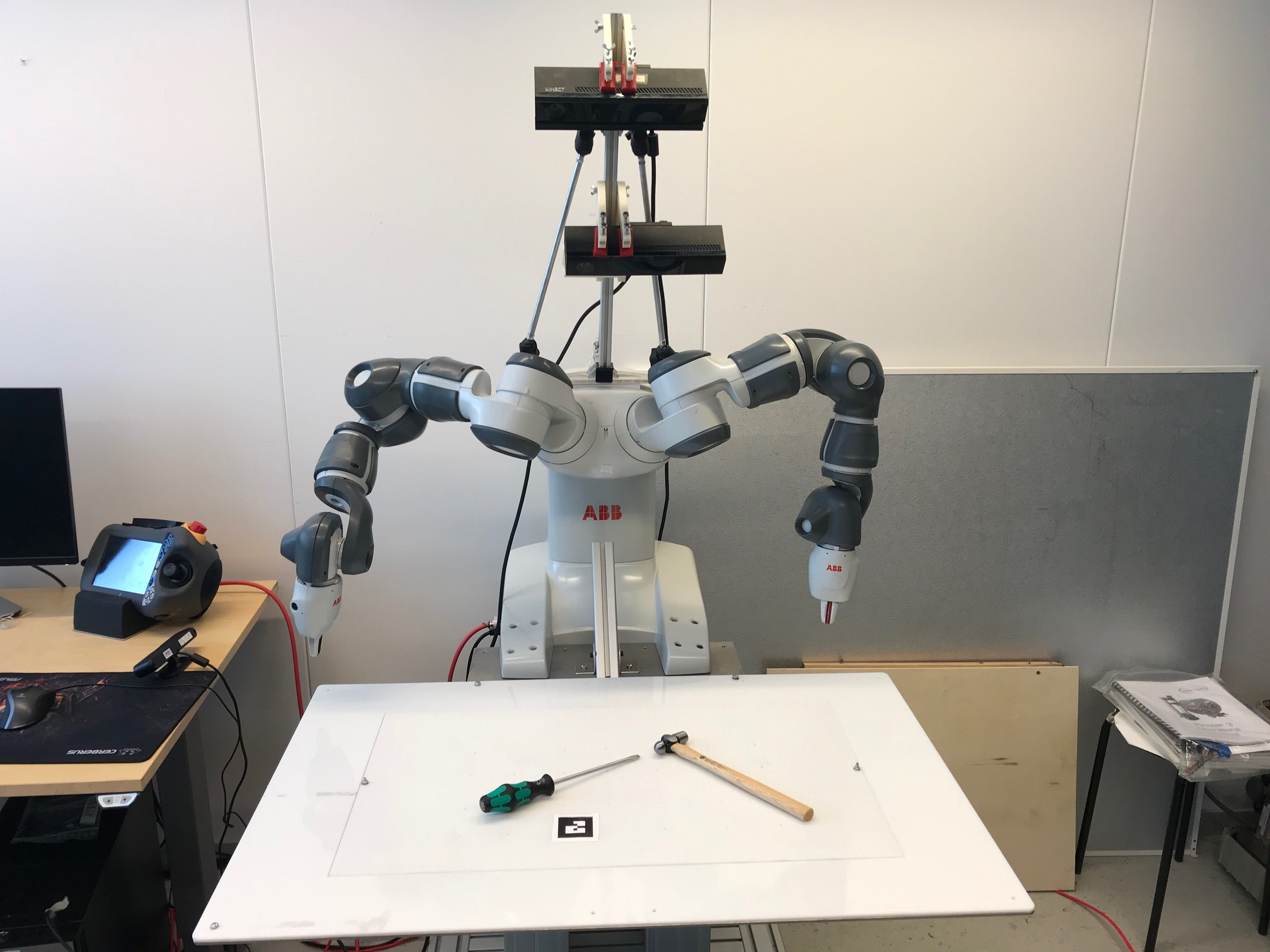}
      \caption{The 2-arm stationary robot platform. }
      \label{fig:robot}
 \end{figure}

The KTH Handtool Dataset\footnote{https://www.nada.kth.se/cas/data/handtool/} is collected for evaluating the object recognition/detection performance of robot vision methods in varying viewpoint, illumination and background settings, all crucial abilities for robot kitting in unconstrained, real-world settings. Instead of having general household objects, the dataset consists of hand tools in order to represent a workshop setting in a factory. It consists of 9 different hand tools for 3 different categories; hammer, plier and screwdriver. The images are collected with a 2-arm stationary robot platform shown in Fig. \ref{fig:robot}. Dataset consists of 3 different illuminations, 2 different cameras (One Kinect camera and one webcam) with different viewpoints and 2 different background settings that correspond to 12 (3x3x2) domains in total. For each hand tool, approximately 40 images with different poses are collected for each camera and domain setting. Table~\ref{tab:dataset} shows example images from different domains. In total, approximately 4500 RGB images are available in the dataset.

\begin{table*}[ht]
\begin{center}
 \caption{Example Images from KTH Handtool Dataset} \label{tab:dataset}
  \begin{tabularx}{0.8\textwidth}{XXXX} 
  \hline  \multirow{2}{*}{Camera Type} & \multicolumn{3}{c}{Illumination} \\ \cline{2-4}
        &  \hspace{5mm} Artificial & \hspace{5mm} Cloudy & \hspace{5mm} Directed  \\ 
      \hline   \vspace{5mm} Kinect & \vspace{0.5mm} \parbox[c]{1em}{\includegraphics[width=1in]{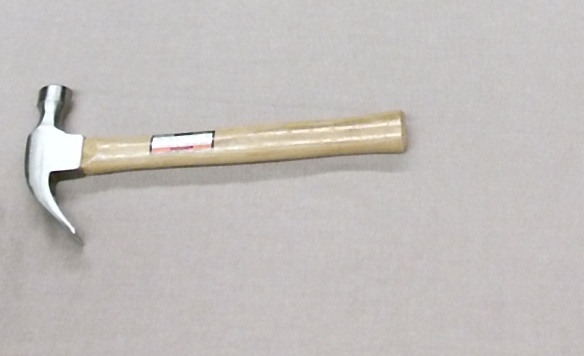}} & 
      \vspace{0.5mm} \parbox[c]{1em}{\includegraphics[width=1in]{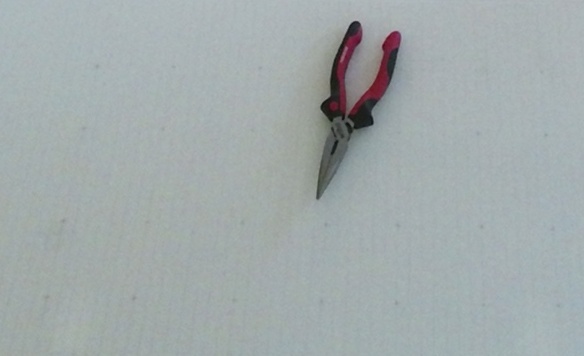}} & 
      \vspace{0.5mm} \parbox[c]{1em}{\includegraphics[width=1in]{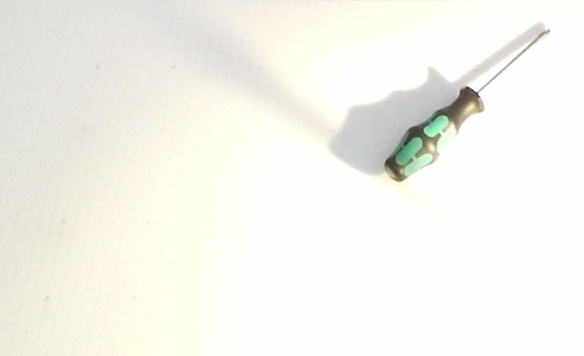}} \\
       \vspace{5mm} Webcam & \vspace{0.5mm} \parbox[c]{1em}{\includegraphics[width=1in]{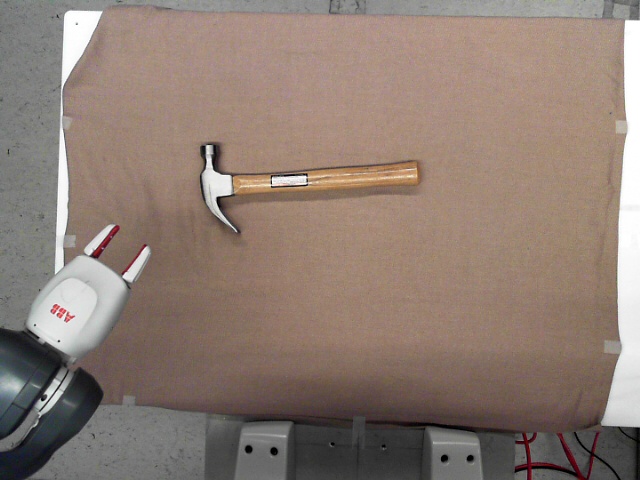}}  & \vspace{0.5mm} \parbox[c]{1em}{\includegraphics[width=1in]{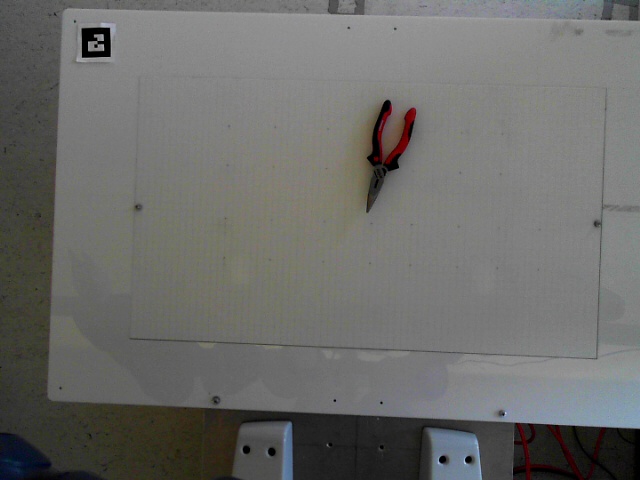}} & \vspace{0.5mm} \parbox[c]{1em}{\includegraphics[width=1in]{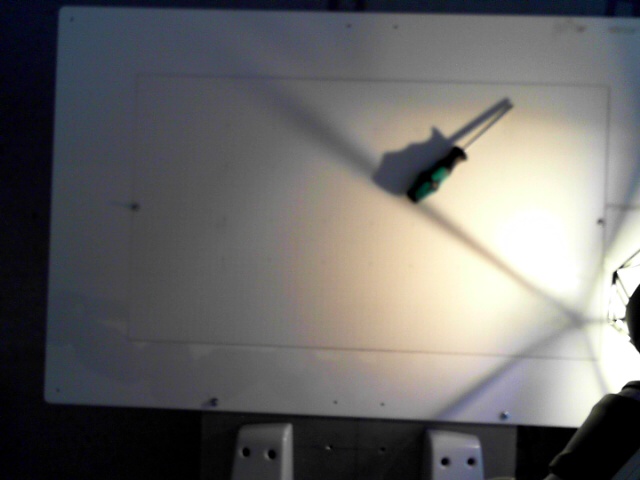}} \vspace{2mm}\\  
       \hline
  \end{tabularx}
\end{center}
\end{table*}


\section{Online Domain Adaptation}
\label{sec:method}
In this section we present our strategy for performing online domain adaptation by exploiting Batch Normalization (BN) \cite{ioffe2015batch}. After giving a formal definition of Domain Adaptation (section \ref{DA}), we recall the basic principles of BN and discuss how it can be exploited into a neural architecture for reducing the domain shift (section \ref{DA-BN}). Then, we describe our approach for online adaptation of a deep model to novel domains (section \ref{ONDA}). 

\subsection{Domain Adaptation}
\label{DA}
Suppose we collected a set of images using a robotic platform with the aim of training a robot vision model with it. Since the image collection has been acquired in the real world, the resulting model will be biased towards the particular conditions (\eg illumination, environmental) under which the images have been acquired. Because of this, if we employ such a system and the current working conditions are different from those of the training set, the performances of the model will degrade due to the presence of a substantial \textit{shift} between the training and test data. In this situation, to increase the generalization capabilities of the robot we can remove the acquisition bias either by collecting more training data in a large variety of conditions, which is extremely expensive, or by developing algorithms  able to bridge the gap between the training and test data, aligning the original model to the novel scenario. The latter is the goal of domain adaptation.


Formally, we assume to have a source domain $\mathcal{S}=\{I^s_i,y^s_i\}_{i=1}^N$, where $I^s_i$ is an image and $y^s_i \in \{1,\cdots,C\}$ the associated semantic label. Together with the domain $\mathcal{S}$, at training time we assume to have collected a set of images, even unlabeled, of our target domain $\mathcal{T}=\{I^t_1,\cdots,I^t_M\}$. The aim of DA algorithms is to build a deep model $f_\theta$, with $\theta$ denoting the network parameters, able to correctly classify images of the target domain $\mathcal{T}$ by exploiting the labeled data provided for $\mathcal{S}$. In the standard scenario, the set of semantic labels is shared between $\mathcal{S}$ and $\mathcal{T}$.

As discussed in section \ref{sec:related-work}, one of the most successful stream of recent works has addressed DA through BN. Next section summarizes this approach, that will give us the  fundamental tools for our ONline DA (ONDA) algorithm.

\subsection{Domain Adaptation with Batch Normalization}
\label{DA-BN}
BN \cite{ioffe2015batch} is a common strategy for avoiding internal covariate shift within deep learning architectures. It works by normalizing the input features to a fixed, target distribution, \ie a standard Gaussian distribution. Formally, let us denote with $x^{l,k}$ the activations of the $k_{th}$ channel of a layer $l$. In order to perform the normalization, the BN layer requires to compute the mean $\mu_{\mathcal{X}}^{l,k}$ and standard deviation $\sigma_{\mathcal{X}}^{l,k}$ over the training set $\mathcal{X}$ for the activations $x^{l,k}$. Since the formulation is layer and channel independent, in the following we will remove the superscript $l,k$ for sake of clarity. The normalization is performed as follows:
\begin{equation}
\label{eq:bn}
 \hat{x}=\gamma\frac{x-\mu_{\mathcal{X}}}{\sqrt{\sigma_{\mathcal{X}}^2 + \epsilon}} + \beta\,,
\end{equation}
where $\gamma$ is a scale factor and $\beta$ is a bias term, while $\epsilon$ is a constant introduced for numerical stability. 

Since the optimization of the network is usually performed using mini-batches, the statistics of a mini-batch are used as a local estimate of the true BN statistics $\{\mu_\mathcal{X},\sigma^2_{\mathcal{X}}\}$. 
Given a batch $\mathcal{B}$ with $n_b$ samples, the approximate statistics are computed as:
\begin{align*}
\label{eq:bn-update}
{\mu}_{\mathcal{B}} = \frac{1}{n_b}\sum_{i=1}^{n_b} x_i \ \ \ \ \ \ {\sigma}_{\mathcal{B}}^2=\frac{1}{n_b}\sum_{i=1}^{n_b} 
(x_i-{\mu}_{\mathcal{B}})^2
\end{align*} 
where $x_i$ denotes the activations of the $i_{th}$ sample in the mini-batch. The above statistics are exploited to progressively update the global estimate $\{\mu_\mathcal{X},\sigma^2_{\mathcal{X}}\}$. 

\begin{figure}[t]
\centering
\includegraphics[width=1\columnwidth]{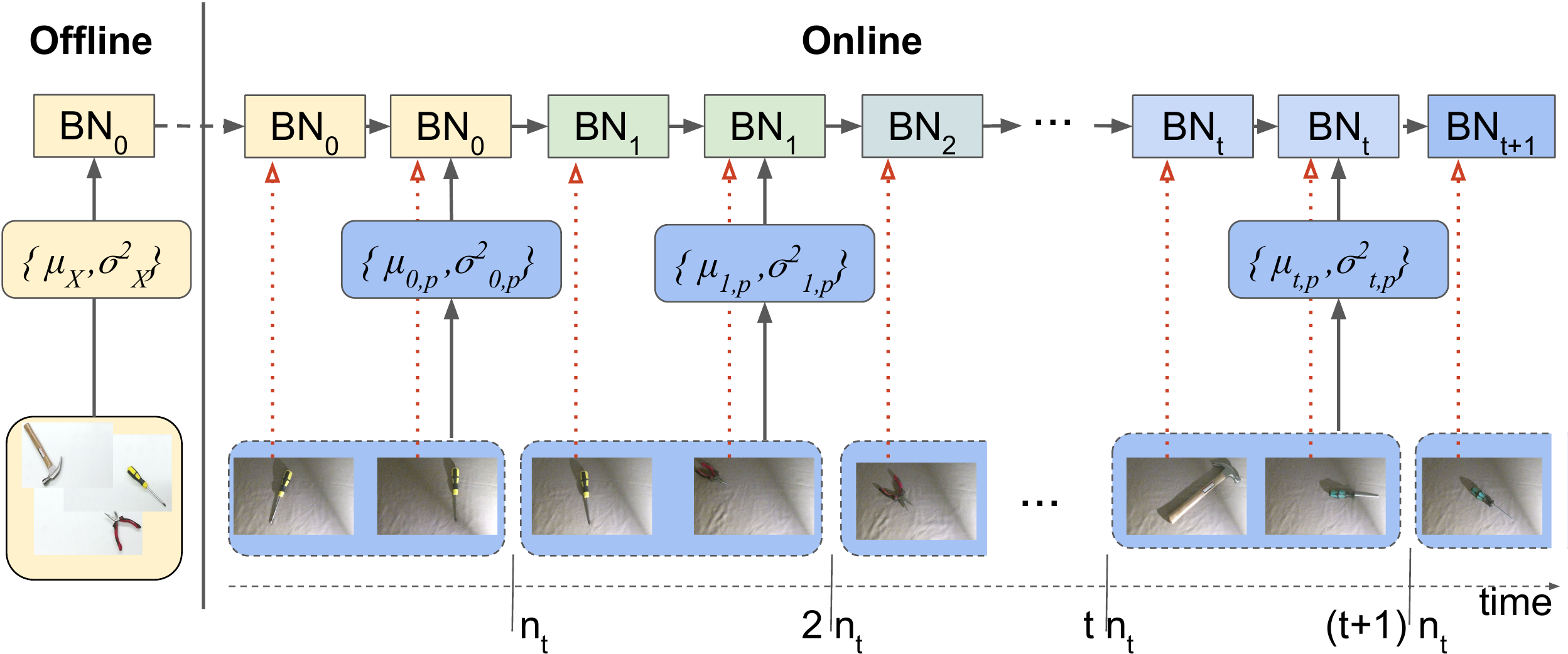}
    \caption{The statistics of the BN layers are initialized offline, by training the network on the images of the source domain. At deployment time, the input frames are processed using the global estimate of the statistics (red lines). However the robot collects each $n_t$ input frames to compute partial BN statistics, using these estimated values to gradually update the BN statistics for the current scenario. } 
   \label{fig:method}
\end{figure}

Recent works \cite{li1603revisiting,carlucci2017just,carlucci2017autodial} have shown how BN layers can be exploited to perform domain adaptation in a traditional batch setting. The main idea behind these works is to create a deep architecture with two parallel branches, one for the source and the other for the target domain. The two branches share the same parameters but embed different domain-specific BN layers. These layers compute different statistics for the source and the target domains, resulting into domain-specific normalizations. In other words, the domain-specific BN layers allow the distributions of features of different domains to be aligned to the same reference distribution, achieving the desired domain adaptation effect. 

\subsection{ONDA: ONline Domain Adaptation}
\label{ONDA}
In this paper we adopt the same idea proposed in \cite{li1603revisiting,carlucci2017just,carlucci2017autodial} but we consider an online setting. Instead of having a fixed target set available during training, we propose to exploit the stream of data acquired while the robot is acting in the environment and continuously update the BN statistics. In this way, we can gradually adapt the deep network to a novel scenario. 


Formally, we consider a different scenario with respect to standard DA algorithms. Opposite to traditional domain adaptation in a batch setting, during training we have only access to the source domain $\mathcal{S}$ and we do not have any data or prior information about the target domain $\mathcal{T}$, apart from the set of semantic labels which is assumed to be shared. When the robot is active, the current working conditions will compose the target domain and we will have access to the automatically acquired sequence of images  $\mathcal{T}=\{I^t_1,\cdots,I^t_{T}\}$. In this scenario, in order to adapt the network parameters $\theta$ to this novel domain, we must exploit the incoming test images collected by the robot on the fly.

If the network contains BN layers, following the idea of previous works \cite{li1603revisiting,carlucci2017just,carlucci2017autodial}, we can perform the adaptation by simply updating the BN statistics with the incoming images of the novel domain. Specifically, we start by training the network on the source domain $\mathcal{S}$, initializing the BN statistics at time $t=0$ as $\{\mu_0,\sigma^2_0\}=\{\mu_\mathcal{S},\sigma^2_\mathcal{S}\}$. 
Assuming that the set of network parameters $\theta$ are shared between the source and target domain except for the BN statistics, we can adapt the network classifier $f_\theta$ by updating the BN statistics with the estimates computed from the sequence $\mathcal{T}$. 
Defining as $n_t$ the number of target images to use for updating online the BN statistics, we can compute a partial estimate $\{\hat{\mu}_t,\hat{\sigma}^2_t\}$ of the BN statistics as: 
\begin{align*}
\hat{\mu_{t}} = \frac{1}{n_t}\sum_{i=1}^{n_t} x_i \ \ \ \ \hat{\sigma_{t}}^2 =\frac{1}{n_t}\sum_{i=1}^{n_t} 
(x_i-\hat{\mu}_{t})^2 
\end{align*} 
The global statistics at time $t$ can be updated as follows:
\begin{align*}
\sigma_{t}^2&=(1-\alpha)\sigma_{t-1}^2+\alpha \frac{n_t}{n_t-1}\hat{\sigma}_{t}^2\\
\mu_{t} &= (1-\alpha) \mu_{t-1} +\alpha\hat{\mu}_{t}
\end{align*}  
where $\alpha$ is the hyper-parameter regulating the decay of the moving average.

The above formulation achieves a similar adaptation effect of the methods \cite{li1603revisiting,carlucci2017just,carlucci2017autodial} but with three main advantages. First, no samples of the target domain, neither labeled nor unlabeled, are used during training. Thus, no further data acquisition and annotation efforts are required. Second, since we do not exploit target data for training, contrary to standard DA algorithms, we have no bias towards a particular target domain. 
Third, since the adaptation process is online, the model can adapt itself to multiple sequential changes of the working conditions, being able to tackle unexpected environmental variations (\eg sudden illumination changes).

The reader might wonder if other possible choices may be considered for initializing $\{\mu_0,\sigma^2_0\}$, such as exploiting a first calibration phase where the robot collects images of the target domain in order to produce a first estimate of the BN statistics. Here we choose to use the statistics estimated on the source domain because 1) we want a model ready to be employed, without requiring any additional preparation at test time; 2) the robot may occur in multiple domains during employment and if a shift occurs (\eg illumination condition changes) our method will automatically adapt the visual model to the novel domain starting from the current estimated statistics: initializing $\{\mu_0,\sigma^2_0\}=\{\mu_\mathcal{S},\sigma^2_\mathcal{S}\}$ allows to check the performance of the algorithm even for multiple sequential shifts and long-term applications. Obviously our method can benefit from a calibration phase or initializations of the statistics closer to the target working conditions: we plan to analyze these aspects in future works.


\section{Experiments}
\label{sec:experiments}
\subsection{Networks and training protocols}
We perform our experiments with the AlexNet \cite{krizhevsky2012imagenet} architecture pre-trained on ImageNet \cite{deng2009imagenet}. We train 3 additional models: a variant of AlexNet with BN, the DA architecture DIAL from \cite{carlucci2017just} and our ONline DA model (ONDA). Following \cite{carlucci2017just}, we add BN layers or its variants after each fully-connected layer. Both the standard AlexNet, AlexNet with BN and DIAL are trained with a batch-size of 128. We implemented \cite{carlucci2017just} by splitting the batch-size between images of source and target domain proportionally to the number of images for each set, as in \cite{carlucci2017just}, without employing the entropy-loss for target images \cite{carlucci2017just,carlucci2017autodial}. We highlight that DIAL is our upper-bound in this case, since it shares the same philosophy of ONDA but with the assumption that images of the target domain are available at training time.

As preprocessing, we rescale all the images in order to ensure a shortest side of 256 pixels, preserving the aspect ratio and subtracting the mean value per channel computed over the ImageNet database. As input to the network we use a random crop of 227$\times$227 at training time, employing a central crop with the same dimensions during test. No additional data augmentation is performed. For all the variants of the architecture, we fine-tune the last layers for 30 epochs with an initial learning rate of 0.001 for \texttt{fc6}, \texttt{fc7} and of 0.01 for the classifier, with a weight decay of 0.0005 and momentum 0.9. The initial learning rates are scaled by a factor of 0.1 after 25 epochs. 

In order to apply our method, we start from the weights of AlexNet with BN, training on the given source domain. Then, we perform one iteration over the target domain, without updating any parameter other than the BN statistics. As a trade-off between stability of the statistics and fastness of adaptation we set $n_t=10$ and $\alpha=0.1$. We will detail the impact of these choices in the following sections.

In all the experiments, we consider the task of object recognition in the \textit{fine-grained} setting, with all the 9 classes considered as classification objective. We report the average accuracy between 5 runs, shuffling the order of the input images in each run of our model.

\subsection{Domain Adaptation results}
\begin{figure*}[t]
  \centering
  \subfloat[Source Domain: Artificial light, Kinect camera and White background]{\includegraphics[width=1\textwidth]{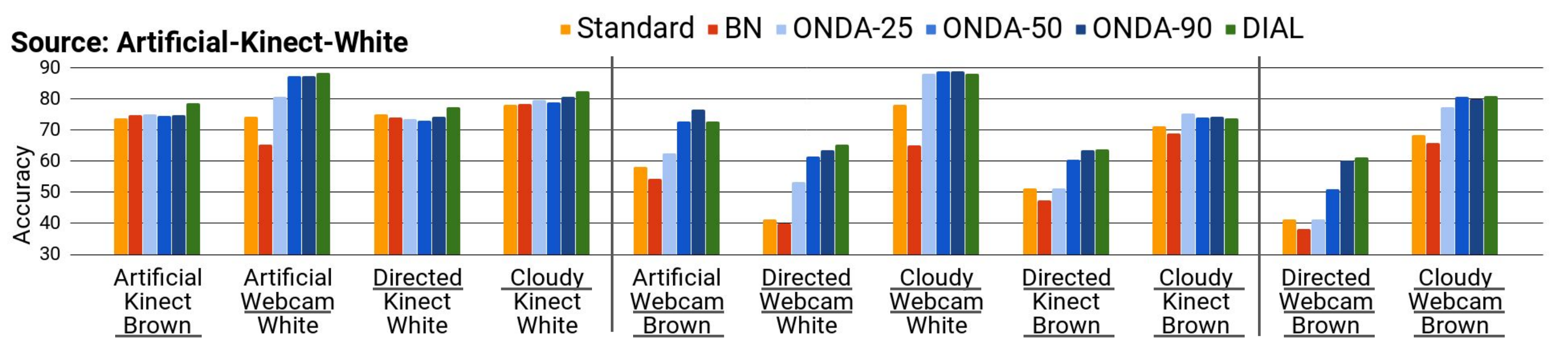}\label{fig:awk-hist}}
    \hfill
  \subfloat[Source Domain: Cloudy light, Webcam camera and Brown background]{\includegraphics[width=1\textwidth]{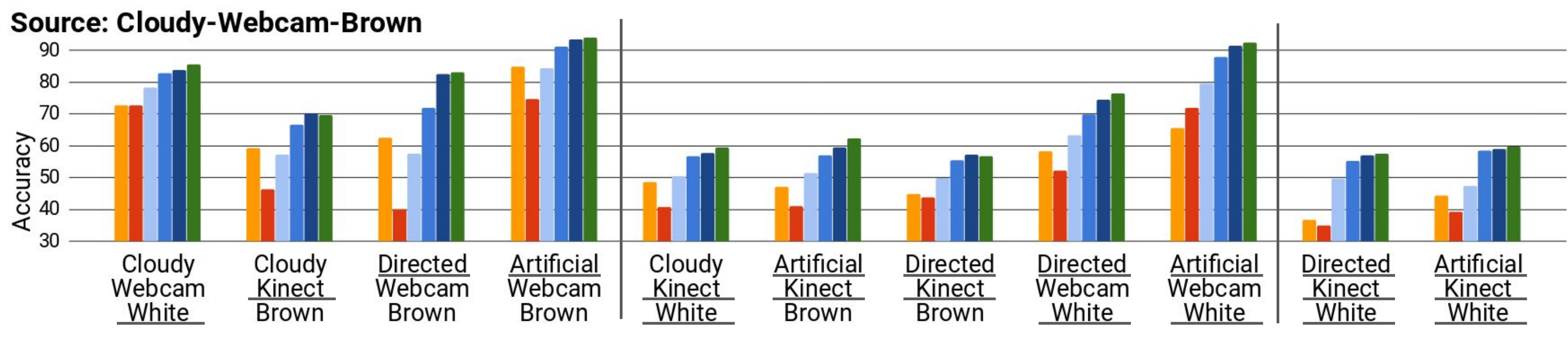}\label{fig:cwb-hist}}
  \caption{Experiments on isolated shifts. The labels of the x-axes denote the conditions of target domain, with the first line indicating the light condition, the second the camera and the third the background. We underlined the changes between the source and target domains. 
  }
  \label{fig:isolated-shifts}
\end{figure*}

In this subsection, we will present the results of our algorithm. 
In order to analyze the particular effect that each possible change may have to the adaptation capabilities of our model, we isolate the sources of shift. To this extent, we consider two sample starting source domains: in the first case (Figure \ref{fig:awk-hist}), the acquisition conditions are artificial light, Kinect camera and white background; in the second case we consider cloudy illumination, webcam and brown background (Figure \ref{fig:cwb-hist}). From these source domains we start by changing only one of the acquisition conditions (left part of the figures) and gradually increasing the number of changes to 2 and 3 conditions (middle and right parts respectively). We report the results for our model after 25\%, 50\% and 90\% of the target data processed. 

As the figures show, 
our model is able to fill the gap between the BN baseline (red bars) and the DA upper bound DIAL (green bars) in almost all settings. 
Only in few cases, where the shift between the performances of BN and DIAL is lower, this does not happen (\ie Figure \ref{fig:awk-hist}, target artificial-Kinect-brown and directed-Kinect-White). In all the other settings the gains are remarkable: considering both figures, the average difference between the performance of BN and ONDA-90 are of 
15\%, 18\% and 20\% for the single, double and triple shift cases respectively. We stress that the gain increases with the amount of shift between the source and target domains, underlying the importance of applying DA adaptation methods in changing environments. 
As expected, the statistics computed in the first stages (\ie ONDA-25) are not always sufficiently representative of the true estimate since they may be still biased by the statistics computed over the source domain. However the estimate becomes more precise as more images of the target domain are processed (\ie ONDA-50 and ONDA-90), gradually covering the gap with the estimate computed by DIAL. The fastness of adaptation and the quality of the estimates depend on the two hyper-parameters $\alpha$ and $n_t$. In the next subsection we will analyze their impact to the final performances of the algorithm.

\subsection{Ablation study}
In this subsection we analyze the impact of the two hyper-parameters, the update frequency $n_t$ and the decay $\alpha$, on the number of images needed by ONDA to estimate the statistics for the target domain. We use a sample scenario of Figure \ref{fig:cwb-hist}, where cloudy illumination, webcam camera and brown background are the source domain conditions and artificial light, Kinect camera and white background are the target domain ones. In the first experiment, we fix $n_t$ to 10, varying the value of $\alpha$. We start by a single pre-trained model of AlexNet with BN repeating the experiments for 5 runs, shuffling the order of the input data, and reporting the average accuracy for each update step.

\begin{figure}[t]
\centering
\includegraphics[width=\columnwidth]{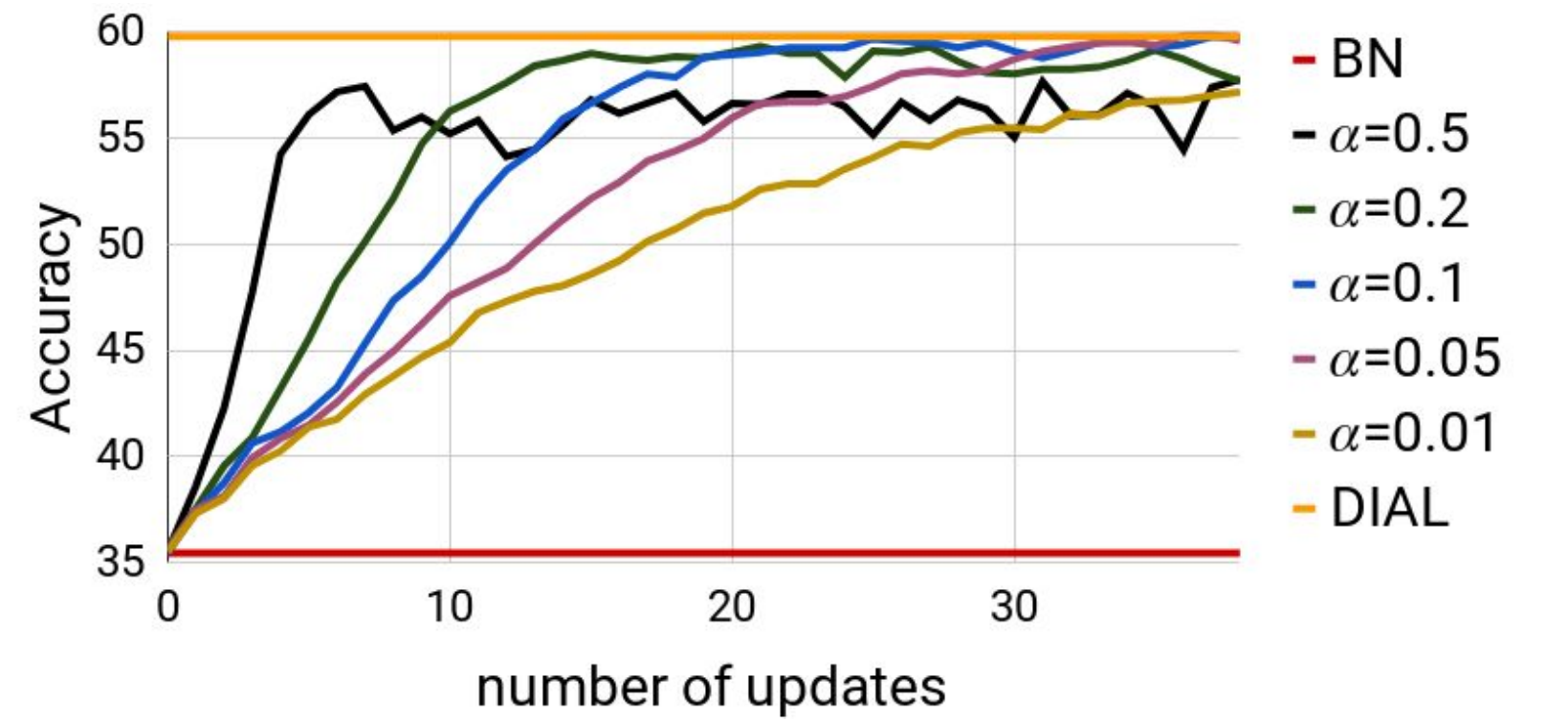}
    \caption{Accuracy vs number of updates of ONDA for different values of $\alpha$ fixing $n_t=10$ in a sample scenario. The red line denotes the BN lower bound of the starting model, while the yellow line the DIAL upper bound.} 
   \label{fig:graph-alpha}
\end{figure}

Results are shown in Figure \ref{fig:graph-alpha}: increasing the value of $\alpha$ to 0.2 (green line) or 0.5 (black line) allows the model to achieve a faster adaptation to the target conditions, with the drawback of a noisier estimation of the statistics. Thus, increasing $\alpha$ leads to an unstable convergence of the performance. On the other hand, choosing too low values of $\alpha$ (\eg 0.05 or 0.01, purple and gold lines respectively) allows a more stable convergence of the model, but with the drawback of slower adaptation to the novel conditions. 

\begin{figure}[t]
\centering
3\includegraphics[width=\columnwidth]{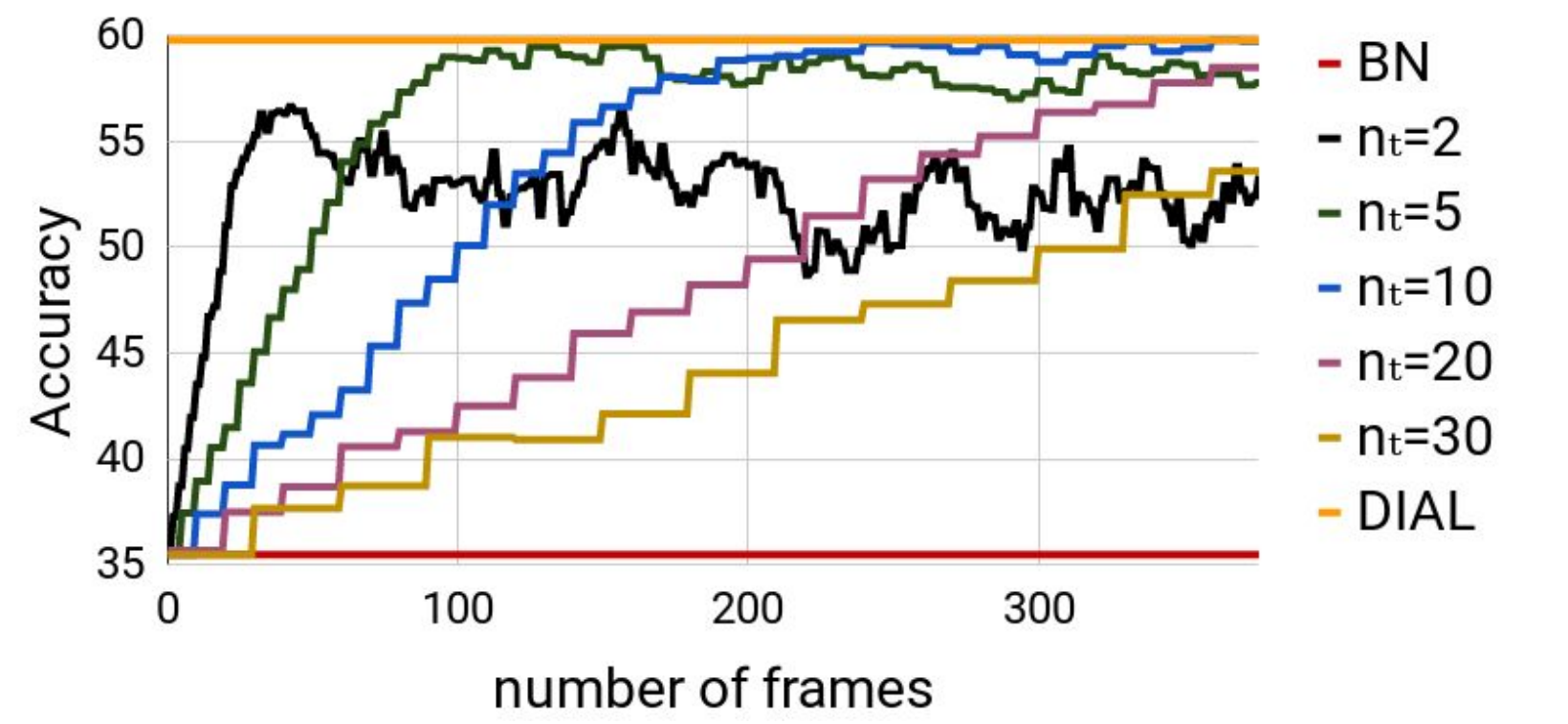}
    \caption{Accuracy vs number of frames processed of ONDA for different values of $n_t$ fixing $\alpha=0.1$ in a sample scenario. The red line denotes the BN lower bound of the starting model, while the yellow line the DIAL upper bound.} 
   \label{fig:graph-nt}
\end{figure}

Regarding the hyper-parameter $n_t$, we follow the same protocol of the first experiment, fixing $\alpha$ to 0.1 and varying the number of images collected before updating the statistics, $n_t$, reporting how the accuracy changes with respect to the number of frames processed. As Figure \ref{fig:graph-nt} shows, low values of $n_t$ (\eg $n_t=2$) allows a faster adaptation, due to the higher update frequency, but at the price of a noisier estimation of the statistics, which is harmful to the final accuracy achieved by the model. At the same time, high values of $n_t$ (\eg 20, 30) allow for a more precise estimate of the statistics, highlighted by the smoothness of the respective lines in the graph, with the drawback of a lower speed of adaptation to the novel domain, caused by the lower update frequency.

The speed of adaptation and the final quality of the BN statistics is obviously a consequence of the values chosen for both hyper-parameters. Obviously $\alpha$ and $n_t$ are not independent from each other: for a lower $n_t$ a lower $\alpha$ should be selected in order to preserve the final performance of the algorithm and conversely for a higher $n_t$, a higher $\alpha$ will allow a faster adaptation of the model. As a trade-off between fast adaptation and good results, we found experimentally that choosing $n_t=\{5,10,20\}$ and $\alpha=\{0.05,0.1\}$ worked well for both short and long term experiments. 

\section{Conclusions}
\label{sec:conclusions}
In this work, we presented a novel dataset for addressing the kitting task in robotics. The dataset takes into account multiple variations of acquisition conditions such as camera, illumination and background changes which may occur during the robot employment. This dataset is intended for testing the robustness of robot vision algorithms to changing environments, providing a novel benchmark for assessing the robustness of robot vision systems.

Together with the dataset, we proposed an algorithm which is able to perform online adaptation of deep models to unseen scenario. The algorithm, based on the update of the statistics of batch-normalization layers, is able to continuously adapt the model to the current environmental conditions of the robot, providing more robustness to unexpected working conditions. Experiments on the newly proposed dataset, confirm the ability of our algorithm to fill the gap between a standard architecture and its domain adapted counterpart without requiring any additional target data during training.

As future works, we plan to enlarge the dataset, including more source of variations and more objects. We further plan to provide a deeper analysis of our algorithm with more architectures, as well as exploring possible extensions which could exploit knowledge coming from previously met scenarios.

\bibliographystyle{IEEEtran}
\bibliography{IEEEabrv,root}

\end{document}